\documentclass[runningheads]{llncs}

\usepackage[T1]{fontenc}
\usepackage{graphicx}
\usepackage{algorithm}

\usepackage{algpseudocode}
\usepackage{amsmath}
\usepackage{enumitem}
\usepackage{xcolor}
\usepackage{mdframed}
\usepackage{url}
\usepackage{tikz}
\usepackage{pgfplots}
\pgfplotsset{compat=1.18}
\usepgfplotslibrary{statistics}
\usepgfplotslibrary{groupplots}
\usepackage{listings}
\usepackage{xcolor}
\usepackage[most]{tcolorbox}
\lstdefinestyle{promptstyle}{
  basicstyle=\small\ttfamily,
  breaklines=true,
  breakindent=0pt,
  breakatwhitespace=false,
  columns=fullflexible,
  keepspaces=true,
  showstringspaces=false,
  frame=single,
  numbers=none,
  xleftmargin=0pt,
  aboveskip=4pt,
  belowskip=4pt,
  backgroundcolor=\color{gray!5}
}
\newtcblisting{promptbox}{
  listing only,
  breakable,
  colback=gray!5,
  colframe=gray!60,
  boxrule=0.4pt,
  arc=1mm,
  listing options={
    basicstyle=\ttfamily\small,
    breaklines=true
  }
}
\lstdefinelanguage{FunctionDefinition}{
  keywords={Function, Remark, Input, Output},
  sensitive=true,
  comment=[l]{//},
  morecomment=[s]{/*}{*/},
  morestring=[b]"
}
\newtcblisting{functionlisting1}{
  breakable,
  enhanced,
  listing only,
  colback=gray!10,
  colframe=gray!60,
  boxrule=0.4pt,
  arc=2mm,
  listing options={
    language=FunctionDefinition,
    basicstyle=\normalfont\ttfamily\small,
    keywordstyle=\bfseries,
    identifierstyle=\itshape,
    commentstyle=\itshape,
    stringstyle=\normalfont,
    breaklines=true
  },
  use counter=lstlisting,
  list inside=lstlisting
}
\newtcblisting{functionlisting}{
  breakable,
  enhanced,
  listing only,
  colback=gray!10,
  colframe=gray!60,
  boxrule=0.4pt,
  arc=2mm,
  listing options={
    basicstyle=\ttfamily\small,
    breaklines=true
  },
  use counter=lstlisting,
  list inside=lstlisting
}
\usepackage{subcaption}

\definecolor{RoyalBlue}{RGB}{31,119,180}
\pgfplotsset{
  colormap={bluewhiteRed}{
    rgb(0pt)=(0.1,0.4,0.8)  
    rgb(128pt)=(1,1,1)      
    rgb(255pt)=(0.8,0,0)    
  }
}
\usetikzlibrary{arrows.meta,positioning}
\tikzset{
  mybox/.style={
    draw,
    rounded corners=2pt,
    line width=0.6pt,
    text width=#1,
    align=left,
    inner sep=6pt,
    font=\footnotesize,
  }
}

\algrenewcommand{\algorithmiccomment}[1]{\hspace{0.2cm}// #1}
\newcommand{\Linecomment}[1]{\Statex \hspace{\algorithmicindent} \textit{// #1}}
\algrenewcommand\algorithmicrequire{\textbf{Input:}}
\algrenewcommand\algorithmicensure{\textbf{Output:}}

\newcommand{\sparql}{\texttt{SPARQL}}

\newcommand{\call}[2]{{$\textsc {#1}$}(#2)}

\newcommand{\llm}{\textsc{LLM}}
\newcommand{\graphrag}{Graph-RAG}

\newcommand{\gptresponse}{\textsc{LlmResponse}}

\newcommand{\rag}{\textsc{OntologyBasedRetrieval}}

\newcommand{\stoptoken}{\texttt{<STOP>}}
\newcommand{\varclasses}{\texttt{classes}}

\newcommand{\varretrievedinfo}{\texttt{retrieved\_info}}
\newcommand{\vargptresponse}{\texttt{gpt\_response}}
\newcommand{\varnodedict}{\texttt{node\_dict}}

\newcommand{\varretrievednodes}{\texttt{retrieved\_nodes}}
\newcommand{\varusermessage}{\texttt{user\_m}}

\newcommand{\return}[1]{\textbf{return } #1}
\newcommand{\none}{\texttt{None}}

\algnewcommand{\IfThen}[2]{
\State \algorithmicif\ #1\ \algorithmicthen\ #2 }
\algnewcommand{\IfThenElse}[3]{
\State \algorithmicif\ #1\ \algorithmicthen\ #2\ \algorithmicelse\ #3}

\begin{document}

\title{Using Large Language Models and Knowledge Graphs to Improve the Interpretability of Machine Learning Models in Manufacturing}
\titlerunning{Improving the Interpretability of Machine Learning in Manufacturing}

\author{
  Thomas Bayer\orcidID{0009-0007-4373-7933} \and
  Alexander Lohr\orcidID{0009-0006-1612-6793} \and
  Sarah Weiß\orcidID{0009-0004-1811-5065} \and
  Bernd Michelberger\orcidID{0009-0003-0058-507X} \and
Wolfram Höpken\orcidID{0000-0002-4175-1295}}

\authorrunning{T. Bayer et al.}

\institute{University of Applied Sciences Ravensburg-Weingarten, Germany\\
\email{\{firstname.lastname\}@rwu.de}}

\maketitle

\begin{abstract}
  Explaining Machine Learning (ML) results in a transparent and user-friendly manner remains a challenging task of Explainable Artificial Intelligence (XAI). In this paper, we present a method to enhance the interpretability of ML models by using a Knowledge Graph (KG). We store domain-specific data along with ML results and their corresponding explanations, establishing a structured connection between domain knowledge and ML insights. To make these insights accessible to users, we designed a selective retrieval method in which relevant triplets are extracted from the KG and processed by a Large Language Model (LLM) to generate user-friendly explanations of ML results. We evaluated our method in a manufacturing environment using the XAI Question Bank. Beyond standard questions, we introduce more complex, tailored questions that highlight the strengths of our approach. We evaluated 33 questions, analyzing responses using quantitative metrics such as accuracy and consistency, as well as qualitative ones such as clarity and usefulness. Our contribution is both theoretical and practical: from a theoretical perspective, we present a novel approach for effectively enabling LLMs to dynamically access a KG in order to improve the explainability of ML results. From a practical perspective, we provide empirical evidence showing that such explanations can be successfully applied in real-world manufacturing environments, supporting better decision-making in manufacturing processes.
\end{abstract}

\section{Introduction}
Rapid advances in Machine Learning (ML) have led to significant improvements in various domains, including manufacturing, where ML supports tasks such as predictive maintenance and process automation~\cite{BENHANIFIA2025200501,hoepken2025KIDZ,SCHUCHTER202561}. However, one of the main remaining challenges is the interpretability of ML models. In many cases, users struggle to understand how specific ML results are generated, making it difficult for them to trust and act on ML outcomes~\cite{doshivelez2017rigorousscienceinterpretablemachine}.

To address this issue, Explainable Artificial Intelligence (XAI) methods aim to explain ML results in a transparent and user-friendly manner~\cite{8466590}. However, traditional XAI methods often present explanations in ways that are itself difficult for users to understand, as they rely on technical visualizations or complex mathematical concepts. For instance, feature importances are often represented using Shapley values, which are mathematically precise but challenging for non-experts to interpret. In addition, a significant number of XAI methods usually focus only on the ML model itself, without integrating domain-specific knowledge that could help contextualize the results~\cite{9429985}. In manufacturing, where complex processes and interdependencies exist, explanations need to go beyond the internals of ML models and should integrate relevant background knowledge.

In this paper, we present a method to enhance the interpretability of ML models by using a Knowledge Graph (KG). The latter allows to store domain-specific knowledge along with ML results, creating a structured link between data, models, and insights. To make these insights accessible to users, we adopt a Graph-based Retrieval-Augmented Generation (Graph-RAG) approach~\cite{edge2025localglobalgraphrag}, in which relevant subgraphs or triples are selectively retrieved from the KG by a Large Language Model (LLM) in a multiturn dialog. The result is provided as structured context to an LLM for explanation generation. This enables the generation of user-friendly explanations that provide not only transparency but also actionable insights for decision-making in manufacturing processes.

We evaluated our method using the XAI Question Bank \cite{liao2020questioning}. Our evaluation focuses on both quantitative metrics, such as accuracy and consistency, and qualitative ones, such as clarity and usefulness. The results provide insights into how well our method improves explainability in a real-world scenario.

The remainder of this paper is structured as follows: Section 2 provides an overview of related work, covering KGs and LLMs. Section 3 describes our methodology in detail. Section 4 presents our evaluation results, followed by a discussion in Section 5, highlighting key findings, strengths, and limitations. Finally, Section 6 concludes the paper and discusses future research challenges.

\section{Related Work}
The use of an \llm \ to query a knowledge base (KB) in order to answer a question instead of relying on learned facts during pretraining/finetuning leverages information retrieval, since it provides up to date information and avoids false answers because of hallucination. As a prerequisite, the available knowledge must be made accessible through a structured knowledge base (KB) and the \llm \ must be able to construct queries to the KB from natural language.

\paragraph{Ontologies and Knowledge graphs.}
ML-Schema, introduced by Esteves et al. \cite{mlschema:201610}, constitutes a specific ontology for describing ML models, datasets, and experiments, thereby enabling interoperability, reproducibility, and transparency across machine learning workflows. Its applicability at scale has been demonstrated by Kumar et al. \cite{kumar2019ontologies} in domain-specific KGs for industrial and Industry~4.0 settings. Complementary machine-learning ontologies such as MLSea, proposed by Dasoulas et al., further enrich this landscape by introducing semantic layers that improve the discoverability and interlinking of ML artefacts \cite{dasoulas2024mlsea}. Several studies analyze and evaluate ML-Schema in practice. Garcia et al.\ review ontologies for data mining and ML and highlight their role in improving interoperability and reuse \cite{garcia2023ontologies}. Khan et al.\
empirically demonstrate that standardized semantic representations substantially improve interoperability and collaboration between data mining tools \cite{khan2023evaluating}. Nguyen et al.\ and Wilson et al.\ further show that ML-Schema effectively supports collaborative mining scenarios and integrates well into existing machine learning platforms \cite{nguyen2023advancing,wilson2023integration}.

In our work, we build on MLSchema and present a novel and refined ontology for representing complete ML processes and models, including explanations and results of XAI techniques like Shapely values.

\paragraph{Knowledge graphs and LLMs.}
Recent studies explore the combination of KGs and \llm s to improve interpretability and access to structured knowledge.
To address the limited and static knowledge of standalone \llm s, Xiao et al.\ present Ontology-Based Data Access (OBDA) as a framework for accessing heterogeneous data sources via an ontology layer \cite{ijcai2018p777}. Building on this foundation, Liang et al.\ propose a natural-language-to-\sparql \ translation approach for querying knowledge graphs \cite{liang2021querying}. Pan et al. \cite{Pan2024}\ identify three complementary paradigms for integrating knowledge graphs and \llm s, namely KG-enhanced \llm s, \llm-augmented KGs, and synergistic \llm–KG frameworks. Representative systems such as LLM4QA by Lan et al.\ and SGPT by Rony et al.\ demonstrate how fine-tuned \llm s can generate \sparql \ queries from natural language questions, improving interpretability and retrieval efficiency in KG question answering \cite{Lan2024LLM4QA,rony2022sgpt}. Kaplan et al.\ illustrate the growing use of combined \llm \ and knowledge graph approaches in application domains such as software architecture research \cite{KaplanKeimSchneider2024_1000171637}.
Zhu et al. \cite{edge2025localglobalgraphrag} propose \graphrag, which retrieves query-relevant subgraphs from a KG and uses them as structured context for LLMs.

In our study, we present a specific approach for traversing the KG by an \llm \ using multi-turn conversations to answer a user query. We abstain from generation of \sparql \ queries, since these are further sources of error. Instead, our approach uses prompt-tuning, where relevant parts of the KG are provided to the \llm \ incrementally, enabling a flexible in-depth extraction of information.

\section{Methodology}

The presented approach is situated within the XAI category of \textit{post-hoc explainability} as defined by  Arrieta et al. \cite{BARREDOARRIETA202082}, encompassing techniques applied to opaque ML models after training to enhance their interpretability. Specifically, we combine structured knowledge representations through KGs  with \llm s to generate comprehensible, natural language explanations. This methodology aligns with the post-hoc paradigm in that it does not alter the internal workings of the ML model itself, but instead provides explanations by retrieving and reformulating relevant information derived from the model's output and domain-specific knowledge.  Our use of KGs to provide information, as well as to structure and contextualize model outputs, introduces a novel layer of semantic integration, which is then accessed by LLMs capable of retrieving information and answering user queries based thereon.
In accordance with the taxonomy established by Arrieta et al. \cite{BARREDOARRIETA202082}, the proposed method represents a hybridization of \textit{explanation by simplification} and \textit{text-based post-hoc techniques}, aiming to maximize intelligibility for both domain experts and end users.

Recent findings by Ovadia et al. \cite{Ovadia2024} further validate our decision to use \graphrag \ as the foundation of our explanation method. In a comprehensive evaluation comparing fine-tuning and RAG across several knowledge-intensive tasks, the authors show that RAG consistently outperforms fine-tuning — not only in scenarios involving new or domain-specific knowledge but also in revisiting previously seen information. Crucially, \graphrag \ avoids issues such as \textit{catastrophic forgetting} and the need for repeated exposure to the same knowledge in varied forms, which are commonly associated with fine-tuning. Instead, RAG enables the use of an external, structured knowledge source (in our case, a Knowledge Graph) to inject relevant, up-to-date information into the model's reasoning process without altering the underlying model weights. These findings support our design decision to combine selective KG-based multi-turn retrieval with \llm-generated natural language explanations, offering a robust and scalable approach to post-hoc interpretability in real-world manufacturing settings.

Although our system is not designed for factoid question answering, we find conceptual parallels with BYOKG, the zero-shot KGQA framework proposed by Agarwal et al.~\cite{Agarwal2024}. Their method demonstrates that symbolic graph exploration, when paired with large language models, can yield high performance even without training data. While BYOKG synthesizes logical programs for QA tasks, our system instead constructs semantically grounded explanation contexts through guided ontology traversal. This reflects a shared architectural pattern—leveraging KGs for structured access to information and \llm s for semantic interpretation—adapted in our case for post-hoc model explainability.

\subsection{Knowledge Graph Construction}\label{section:knowledge_graph_construction}

\begin{figure}
  \centering
  \includegraphics[width=1\linewidth]{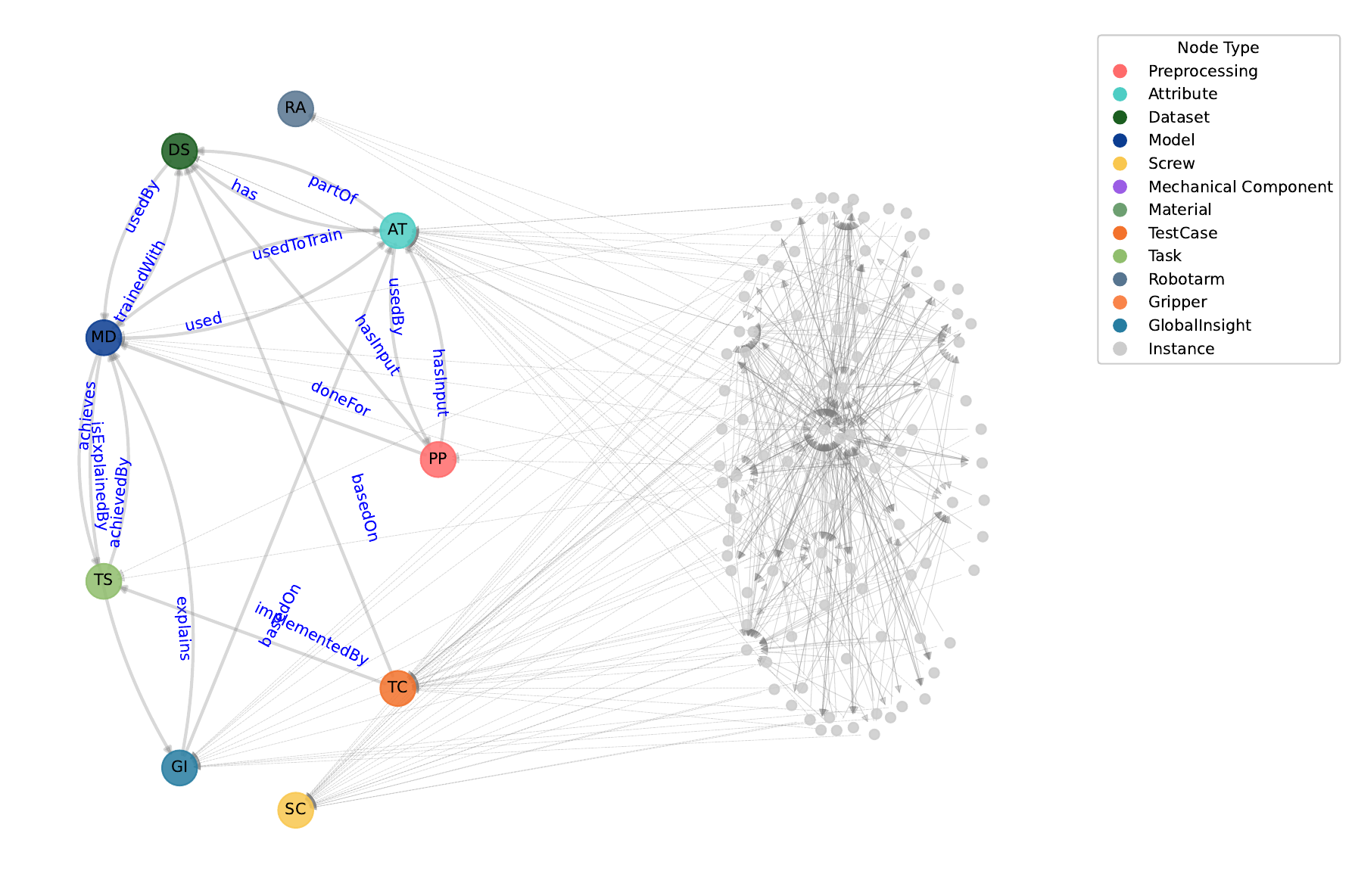}
  \caption{Knowledge Graph: Classes are represented by the colored nodes on the left hand side, instances as gray nodes on the right hand side}
  \label{fig:placeholder}
\end{figure}

\begin{figure}[h]
  \centering
  \begin{tikzpicture}[scale=0.45, transform shape,
      font=\small,
      >=Latex,
      node distance=12mm and 18mm,
      class/.style={
        draw=none,
        fill=red!70!black,
        text=white,
        rounded corners=1.5pt,
        inner sep=6pt,
        align=left,
        minimum width=36mm
      },
      indiv/.style={
        draw=none,
        fill=blue!70!black,
        text=white,
        rounded corners=1.5pt,
        inner sep=6pt,
        align=left,
        minimum width=40mm
      },
      rel/.style={-Latex, line width=0.6pt},
      rlab/.style={midway, fill=white, inner sep=1.5pt, font=\scriptsize}
    ]

    \node[class] (preprocessing) at (-7.2,  3.4) {\textbf{Preprocessing}\\[2pt]
    Describes the\\methods and algorithms\\with which the dataset was\\created};

    \node[class, right=40mm of preprocessing] (dataset) {\textbf{DataSet}\\[2pt]
    A Dataset consisting of\\multiple Rows.};

    \node[class, right=40mm of dataset] (global_insight) {\textbf{Globalsight}\\[2pt]
    A global insight is a\\concrete explanation of a\\model.};

    \node[class, below=70mm of dataset] (model) {\textbf{Model}\\[2pt]
    A model is an\\algorithm trained on data.};

    \node[class, below=70mm of preprocessing] (task) {\textbf{Task}\\[2pt]
    The task that a model\\fulfills};

    \node[indiv,below=20mm of preprocessing] (niryo_preprocessing_a23b) {\textbf{Preprocessing niryo a23b}};
    \node[indiv,below=10mm of niryo_preprocessing_a23b] (niryo_preprocessing_xt77) {\textbf{Preprocessing niryo XT77}};

    \node[indiv] (niryo_ds) at ( -1,  1.5) {\textbf{niryo 2023}};

    \node[indiv, right=20mm of niryo_preprocessing_a23b] (model_a23b) {\textbf{Model a23b}};
    \node[indiv, right=20mm of niryo_preprocessing_xt77] (model_xt77) {\textbf{Model XT77}};

    \node[indiv, above=14mm of task] (screw_placement) {\textbf{ScrewPlacement}};

    \node[indiv, below=20mm of global_insight] (gi_a23b) {\textbf{Global Insights}\\[2pt]A23b};
    \node[indiv, below=10mm of gi_a23b] (gi_xt77) {\textbf{Global Insights}\\[2pt]XT77};

    \node[draw, anchor=south east] at (12,-4.5) {%
      \begin{tabular}{@{}ll@{}}
        \tikz[baseline] \node[
          draw=none,
          fill=red!70!black,
          text=white,
          rounded corners=1.5pt,
          inner sep=6pt,
          align=left,
          minimum width=36mm
        ] (cl) {\textbf{\detokenize{<Name>}}}; & \Large Class \\

        \tikz[baseline] \node[
          draw=none,
          fill=blue!70!black,
          text=white,
          rounded corners=1.5pt,
          inner sep=6pt,
          align=left,
          minimum width=36mm
        ] {\textbf{\detokenize{<Name>}}}; & \Large Individual \\
      \end{tabular}%
    };

    \draw[rel] ([yshift=-3mm]preprocessing.east) -- ++(1.2,0) |- ([yshift=3mm]model.west) node[rlab, near end]{done\_for};
    \draw[rel] (preprocessing.east) -- (dataset.west) node[midway, above]{hasInput};

    \draw[rel] (niryo_preprocessing_a23b.north) -- (preprocessing.south) node[rlab, near start]{type};
    \draw[rel] (niryo_preprocessing_xt77.west) -- ++(-1.2,0) |- (preprocessing.west) node[rlab, near start]{type};
    \draw[rel,dashed] (niryo_preprocessing_a23b.east) -- (niryo_ds.west) node[rlab, near start]{has\_input};
    \draw[rel,dashed] (niryo_preprocessing_xt77.east) -- (niryo_ds.west) node[rlab, near start]{has\_input};
    \draw[rel,dashed] (niryo_preprocessing_a23b.east) -- (model_a23b.west) node[rlab, near start]{done\_for};
    \draw[rel,dashed] (niryo_preprocessing_xt77.east) -- (model_xt77.west) node[rlab, near start]{done\_for};

    \draw[rel] (screw_placement.south) -- (task.north) node[midway, above]{type};
    \draw[rel, <->,dashed] (screw_placement.east) -- (model_a23b.west)
    node[midway, above]{achieves/isAchievedBy};

    \draw[rel, <->,dashed] (screw_placement.east) -- (model_xt77.west)
    node[midway, above]{achieves/isAchievedBy};

    \draw[rel] (model_a23b.south) -- (model.north) node[rlab, near start]{type};
    \draw[rel] (model_xt77.south) -- (model.north) node[rlab, near start]{type};
    \draw[rel] ([xshift=9mm]model.north) -- ([xshift=9mm]dataset.south) node[midway, above]{trainedBy};
    \draw[rel, <->] (model.west) -- (task.east) node[midway, above]{achieves/isAchievedBy};

    \draw[rel] (niryo_ds.north) -- (dataset.south) node[rlab, near start]{type};


    \draw[rel, ->] (global_insight.south) -- (gi_a23b.north)  node[midway, above]{type};

    \draw[rel, ->] (global_insight.east) -- ++(1,0) |- (gi_xt77.east)  node[midway, above]{type};

    \draw[rel] (global_insight.west) -- (model.east) node[midway, above]{explains};

    \draw[rel, ->, dashed] (gi_a23b.west) -- (model_a23b.east)
    node[midway, above]{explains};

    \draw[rel, ->,dashed] (gi_xt77.west) -- (model_xt77.east)
    node[midway, above]{explains};

  \end{tikzpicture}
  \caption{Excerpt from Knowledge Graph}
  \label{fig:ontology}
\end{figure}
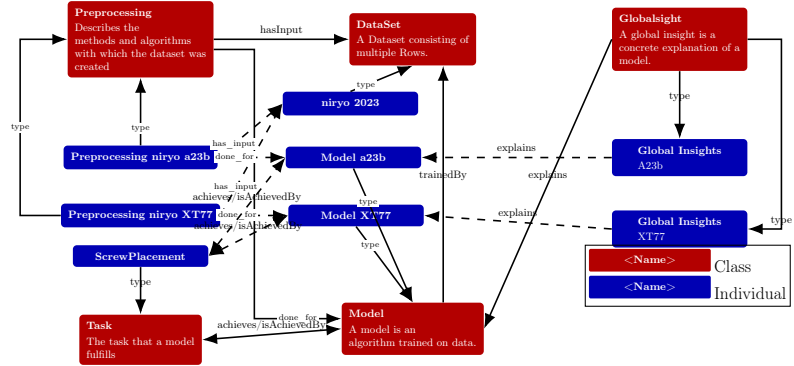

The goal of our KG is to provide information about ML tasks, models, and to facilitate explainability. We have extended ML-Schema \cite{mlschema:201610} in order to support tasks and explanations of models achieving these tasks. The KG contains (as individuals) various ML models, datasets, ML tasks arising in manufacturing, and corresponding explanations. Fig. \ref{fig:placeholder} gives an overview of the KG with classes and their relationships on the left hand side and corresponding instances on the right hand side, while Fig. \ref{fig:ontology} depicts a selected part of the KG, showing more details and concrete instances of the KG.

\subsection{Traversing the Knowledge Graph}

In alignment with the taxonomy outlined by Pan et al.~\cite{Pan2024}, this method utilizes \textit{KG-enhanced \llm s during inference}. In this approach, structured domain knowledge from a knowledge base is not embedded in the \llm \ during pre/post-training or fine-tuning. Instead, it is dynamically retrieved, based on a \textit{Branch-and-Bound} approach, and integrated by the \llm \ using a multi-turn-dialog at the time of query resolution. The retrieved information is then passed to the \llm \ as part of a constructed conversational context, enabling the generation of natural language explanations grounded in both the \llm’s general linguistic capabilities and the domain-specific semantics of the KG.
This architectural choice enables the \llm \ to perform reasoning over current and future data from the KG without the necessity of updating its weights, which is an essential requirement in industrial settings such as manufacturing, where data and process conditions are subject to continuous change.


\begin{figure}[htbp]
  \centering
  \begin{subfigure}{\linewidth}
    \centering
    \footnotesize
    \begin{tcolorbox}[
        colback=blue!10,
        colframe=blue!50!black,
        boxrule=0.5pt,
        arc=2pt,
        width=\linewidth
      ]
      \footnotesize
      \textbf{Question}

      \texttt{List all tasks which are influenced by the dataset niryo september 2024?}
    \end{tcolorbox}
    \vspace{2mm}
  \end{subfigure}

  \begin{subfigure}{\linewidth}
    \centering
    \begin{tikzpicture}[
        >=Stealth,
        every node/.style={font=\scriptsize},
        every edge/.style={opacity=0.3},
      ]
      \definecolor{col_Attribute}{HTML}{4ECDC4}
      \definecolor{col_Dataset}{HTML}{1B5E20}
      \definecolor{col_GlobalInsight}{HTML}{277DA1}
      \definecolor{col_Gripper}{HTML}{F9844A}
      \definecolor{col_Instance}{HTML}{CCCCCC}
      \definecolor{col_Material}{HTML}{6D9F71}
      \definecolor{col_Mechanical_Component}{HTML}{9B5DE5}
      \definecolor{col_Model}{HTML}{0B3D91}
      \definecolor{col_Preprocessing}{HTML}{FF6B6B}
      \definecolor{col_Robotarm}{HTML}{577590}
      \definecolor{col_Screw}{HTML}{F9C74F}
      \definecolor{col_Task}{HTML}{90BE6D}
      \definecolor{col_TestCase}{HTML}{F3722C}
      \tikzset{
        class/.style={
          circle,
          draw,
          fill opacity=0.85,
          minimum size=1.4cm,
          inner sep=2pt
        },
        dataset/.style={class, fill=col_Dataset},
        model/.style={class, fill=col_Model},
        task/.style={class, fill=col_Task}
      }
      \tikzset{
        instance/.style={
          circle,
          draw,
          fill=col_Instance,
          fill opacity=0.85,
          minimum size=1.0cm,
          inner sep=1pt
        }
      }

      \node[dataset] (Dataset) at (-12.0,3) {Dataset};
      \node[model]   (Model)   at (-10,0)  {Model};
      \node[task]    (Task)    at (-12,-3) {Task};

      \node[instance] (ScrewPlacement) at (-5,-3) {ScrewP};
      \node[instance] (model_a23b) at (-2,-3) {a23b};
      \node[instance] (model_xT77) at (-2,-1) {xT77};
      \node[instance] (model_p1b3) at (-2,1) {p1b3};
      \node[instance] (model_qdk1) at (-2,3) {qdk1};
      \node[instance] (niryo_sept) at (-7,3) {niryo};

      \draw[->, line width=0.60pt, solid, draw=gray, bend left=25] (Dataset) to node[midway, fill=white, inner sep=1pt] {usedBy} (Model);
      \draw[->, line width=0.60pt, solid, draw=gray, bend left=25] (Model) to node[midway, fill=white, inner sep=1pt] {trainedWith} (Dataset);

      \draw[->, line width=0.60pt, solid, draw=gray, bend left=25] (Task) to node[midway, fill=white, inner sep=1pt] {achievedBy} (Model);
      \draw[->, line width=0.60pt, solid, draw=gray, bend left=25] (Model) to node[midway, fill=white, inner sep=1pt] {achieves} (Task);

      \draw[->, line width=0.20pt, solid, draw=gray] (ScrewPlacement) to (model_a23b);
      \draw[->, line width=0.20pt, solid, draw=gray] (ScrewPlacement) to (model_xT77);
      \draw[->, line width=0.20pt, solid, draw=gray] (ScrewPlacement) to (model_p1b3);
      \draw[->, line width=0.20pt, solid, draw=gray] (ScrewPlacement) to (model_qdk1);
      \draw[->, line width=0.20pt, dashed, draw=gray] (ScrewPlacement) to (Task);
      \draw[->, line width=0.20pt, solid, draw=gray] (model_a23b) to (niryo_sept);
      \draw[->, line width=0.20pt, solid, draw=gray] (model_a23b) to (ScrewPlacement);
      \draw[->, line width=0.20pt, dashed, draw=gray] (model_a23b) to (Model);
      \draw[->, line width=0.20pt, solid, draw=gray] (model_xT77) to (niryo_sept);
      \draw[->, line width=0.20pt, solid, draw=gray] (model_xT77) to (ScrewPlacement);
      \draw[->, line width=0.20pt, dashed, draw=gray] (model_xT77) to (Model);
      \draw[->, line width=0.20pt, solid, draw=gray] (model_p1b3) to (niryo_sept);
      \draw[->, line width=0.20pt, solid, draw=gray] (model_p1b3) to (ScrewPlacement);
      \draw[->, line width=0.20pt, dashed, draw=gray] (model_p1b3) to (Model);
      \draw[->, line width=0.20pt, solid, draw=gray] (model_qdk1) to (niryo_sept);
      \draw[->, line width=0.20pt, solid, draw=gray] (model_qdk1) to (ScrewPlacement);
      \draw[->, line width=0.20pt, dashed, draw=gray] (model_qdk1) to (Model);
      \draw[->, line width=0.20pt, solid, draw=gray] (niryo_sept) to (model_a23b);
      \draw[->, line width=0.20pt, solid, draw=gray] (niryo_sept) to (model_xT77);
      \draw[->, line width=0.20pt, solid, draw=gray] (niryo_sept) to (model_p1b3);
      \draw[->, line width=0.20pt, solid, draw=gray] (niryo_sept) to (model_qdk1);
      \draw[->, line width=0.20pt, dashed, draw=gray] (niryo_sept) to (Dataset);

      \node[circle, fill=black, minimum size=5pt, inner sep=0pt] (start) at (-13, 0) {start}; 

      \draw[->, line width=2pt, bend left=15, dashed, draw=green] (start) to node[midway, fill=white, inner sep=1pt] {Step 1} (Dataset);
      \draw[->, line width=2pt, bend left=15, dashed, draw=green] (Dataset) to node[midway, fill=white, inner sep=1pt] {Step 2} (niryo_sept);

      \draw[->, line width=2pt, bend left=15, dashed, draw=orange] (niryo_sept) to node[midway, fill=white, inner sep=1pt] {Step 3: 1st iteration} (model_qdk1);
      \draw[->, line width=2pt, bend left=15, dashed, draw=orange] (niryo_sept) to node[midway, fill=white, inner sep=1pt] {} (model_p1b3);
      \draw[->, line width=2pt, bend left=15, dashed, draw=orange] (niryo_sept) to node[midway, fill=white, inner sep=1pt] {} (model_xT77);
      \draw[->, line width=2pt, bend left=15, dashed, draw=orange] (niryo_sept) to node[midway, fill=white, inner sep=1pt] {} (model_a23b);

      \draw[->, line width=2pt, bend right=10, dashed, draw=blue]
      (model_qdk1) to node[midway, sloped,fill=white, inner sep=1pt] {Step 3: 2nd iteration} (ScrewPlacement);

      \draw[->, line width=2pt, bend left=15, dashed, draw=blue]
      (model_p1b3) to node[midway, fill=white, inner sep=1pt] {} (ScrewPlacement);

      \draw[->, line width=2pt, bend left=15, dashed, draw=blue]
      (model_xT77) to node[midway, fill=white, inner sep=1pt] {} (ScrewPlacement);

      \draw[->, line width=2pt, bend left=15, dashed, draw=blue]
      (model_a23b) to node[midway, fill=white, inner sep=1pt] {} (ScrewPlacement);

    \end{tikzpicture}
    \caption{Traversal of the KG: Step 1 identifies the class \textit{Dataset}, step 2 the corresponding instance. Iterations are depicted by orange, resp., blue dashed lines}
    \label{fig:sub1}
  \end{subfigure}

  \vspace{0.5em}

  \begin{subfigure}{\linewidth}
    \centering
    \footnotesize
    \begin{tcolorbox}[
        colback=green!10,
        colframe=green!50!black,
        boxrule=0.5pt,
        arc=2pt,
        width=\linewidth
      ]
      \footnotesize
      \textbf{Answer}

      \texttt{The only task explicitly linked to the *niryo\_dataset\_september\_2024* in the ontology is **ScrewPlacement**. All models that achieve this task rely on the attributes and examples contained in that dataset to learn how to place screws correctly.}
    \end{tcolorbox}

    \vspace{1mm}

    \begin{tcolorbox}[
        colback=red!12,
        colframe=red!50!black,
        boxrule=0.5pt,
        arc=2pt,
        width=\linewidth
      ]
      \footnotesize
      \textbf{Expected Answer}

      \texttt{The task ScrewPlacement is uniquely influenced by the dataset niryo\_september\_2024, as all models capable of performing this task have been trained using the specified dataset.}
    \end{tcolorbox}
    \caption{Answer to the question in Fig. \ref{fig:sub1} generated by our prototype - below the expected answer generated manually}
    \label{fig:sub2}
  \end{subfigure}
  \caption{Schematic example of KG traversal of our \graphrag \ based prototype, where missing task-specific context is retrieved based on the user query}
  \label{fig:graph_traversal}
\end{figure}




This design mirrors Pan et al.’s recommendation for inference-time KG integration, which is characterized by modularity, factual consistency, and adaptability. Unlike approaches that rely on static fine-tuning or pretraining over graph data, our method leverages the flexibility of \graphrag \ to interact with the KG on demand. Moreover, the iterative traversal logic implemented in our retriever — driven by \llm \ responses until a termination criterion is met — exemplifies the kind of dynamic, multi-hop reasoning that Pan et al. highlight as a key advantage of synergistic KG-\llm \ systems.
The multi-turn dialog terminates if either the \llm \ indicates that sufficient information has been retrieved (by responding with the codeword \stoptoken) or no new information has been obtained by querying the KG.
Since the retrieved information of each turn is collected and the knowledge graph is finite, termination of the algorithm is ensured.





For prompting an \llm, we use openai-chat format, cf. \cite{openai_chat_format}, and the function \gptresponse \ (Fig. \ref{fig:llm_call}), which can be implemented for arbitrary \llm s.
\begin{figure}[ht]
  \centering
  \begin{tcolorbox}[
      colback=gray!10,
      colframe=gray!60,
      boxrule=0.4pt,
      arc=1mm
    ]
    \textbf{Function:} $\gptresponse(\textit{user\_message}, \textit{system\_message}, \textit{history})$

    \textbf{Remark:} Function calls API of an LLM, uses OpenAI chat format

    \textbf{Input:}\\
    \hspace*{1.5em}\textit{user\_message}: User-prompt\\
    \hspace*{1.5em}\textit{system\_message}: System message for \llm\\
    \hspace*{1.5em}\textit{history} (optional): previous chat, for multiturn, will be modified\\[0.4em]
    \textbf{Output:}\\
    \hspace*{1.5em}$response$ of LLM API call\\
    \hspace*{1.5em}$history$ object updated accordingly
  \end{tcolorbox}\caption{Function call}\label{fig:llm_call}
\end{figure}


The \graphrag \ workflow integrates the responses of an \llm \ into three steps that structure the retrieval and reasoning process:

\begin{itemize}[label={}, left=30pt] 
  \item[Step 1:] \textit{Identification of the Most Relevant Node Class}: Based on the ontology of the KG, a system message is constructed (see Listing \ref{lst:sds_prompt}). Together with the user query, it forms the complete prompt that is submitted to the \llm \ . The \llm \ then identifies the most relevant ontology classes with respect to the query. If no suitable candidates are returned, a simplified fallback query is issued to determine potential matches.

  \item[Step 2:] \textit{Identification of Starting Node(s)}: Based on Step 1, the \llm \ identifies the ontology node(s) corresponding to instances that best match the query context, serving as entry points for subsequent retrieval.

  \item[Step 3:] \textit{Iterative Ontology Search}: This step consists of two sequential requests. First, the \llm \ receives the identified start node(s) along with structural information about the data, system capabilities, and retrieval objectives. Second, it conducts an iterative search, expanding through the ontology in a Branch-and-Bound-manner until enough information is gathered or no new information is found, cf. \rag
    \\ (Algorithm \ref{alg:iterate_ontology})
\end{itemize}

We only provide the algorithm for Step (3), cf. Algorithm \ref{alg:iterate_ontology} below, since step (1) and (2) are straightforward. An illustration can be found in Fig. \ref{fig:graph_traversal}.
The retrieval algorithm is implemented within a prototype used for the user study in Section~\ref{section:results}. The complete prototype, including prompt templates and our question bank, is publicly available on GitHub\footnote{\label{fn:github_repo}\url{https://github.com/bayerth/kidz_knowledge_graph_rag}}. All evaluations use the OpenAI model \texttt{gpt-4o-2024-11-20}, with temperature 1 and a fixed random seed to ensure reproducibility.


\begin{figure}[h]
\begin{lstlisting}[style=promptstyle]
#System Message
The following structure illustrates the class level of the ontology, which will be used to answer the subsequent questions. The node classes have instances that are not listed here: {ontology_structure}.

#User Message
Only give as an answer a list of classes (following this syntax:
[class1, class2, ...]) which are relevant for this user query {query}
Return only JSON syntax without prefix.
\end{lstlisting}
  \caption{Prompt Template for Initial Step}\label{lst:sds_prompt}
\end{figure}

\begin{algorithm}[t]
  \caption{Information Retrieval with Graph Traversal}
  \label{alg:iterate_ontology}
  \begin{mdframed}[backgroundcolor=lightgray!20, roundcorner=4pt]
    \begin{algorithmic}[1]
      \Require An Ontology object \textit{ontology}, $\llm-history$ of Step 1,2
      \Ensure A list of relevant instances of \textit{ontology} w.r.t. \textit{query}

      \Function{\rag}{ontology, history}
      \Linecomment{Steps 1,2: Identify classes and initial instances of $ontology$ w.r.t. $query$}
      \State $\varclasses, instances$ are initialized in Step 1 and Step 2, $history$ also
      \IfThen {$classes$ is empty}{\return $[\ ],\none,history$} \Comment{No result for $query$}
      \Linecomment{Step 2: Prepare Iteration for Ontology Traversal}
      \State \varnodedict = ontology.get\_nodes($instances$)
      \State \varretrievednodes = \varnodedict.values()
      \Linecomment{Step 3: Start Iteration for Ontology Traversal}
      \Repeat
      \If {$\varretrievednodes == [\ ]$}
      \State  $\varretrievednodes = $\call{execute\_query}{\vargptresponse, \textit{ontology}}
      \EndIf
      \If{\varretrievednodes \ is not empty}
      \For{node in \varretrievednodes}
      \State \varretrievedinfo \texttt{ += } ontology.\text{get\_node\_structure}(\text{node})
      \State update \varnodedict \ with node
      \EndFor
      \Else
      \State $\varretrievedinfo = \texttt{"No instance found"}$
      \EndIf
      \State \varusermessage = "\texttt{Result to your query: \texttt{str(\varretrievedinfo)}}.
      \Statex \hspace{1.5cm} \texttt{If you need more information, use another query,}
      \Statex \hspace{1.5cm} \texttt{otherwise write \stoptoken. Return JSON without prefix.}"
      \State $\vargptresponse, history =$ \call {\gptresponse} {\varusermessage,$system$ \textit{history}}
      \State $\varretrievedinfo = [\ ]$
      \Until {\stoptoken \textbf{ in} $gpt\_response$ or no new info retrieved}
      \State \return $\varnodedict,history$
      \EndFunction
    \end{algorithmic}
  \end{mdframed}
\end{algorithm}

\section{Evaluation and Results}
\label{section:results}

Our prototype is applied to a manufacturing setting where a robotic manipulator places screws into holes at varying angles. The placement success is predicted based on screw geometry and robot-arm attributes. The KG described in Section~\ref{section:knowledge_graph_construction} provides information on tasks, models, hardware, and their relations.

\subsection{Evaluation Methodology}

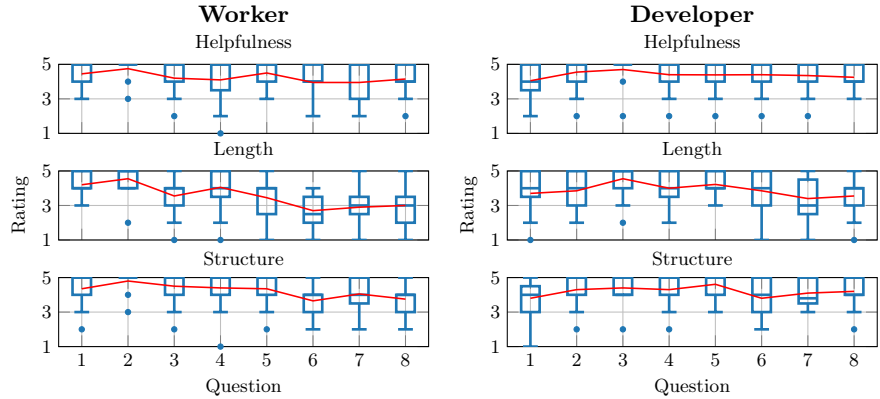
\begin{figure*}[t]
  \centering
  \footnotesize
  \def\myboxlinewidth{1.0pt}
  \def\mylocalwidth{6.5cm}
  \def\mylocalheight{2.5cm}

  \begin{subfigure}{0.48\textwidth}
    \centering
    \begin{tikzpicture}

      \begin{axis}[
          hide axis,
          height=50pt,
          width=\mylocalwidth,
          title={\bfseries Worker},
          title style={yshift=28pt, font=\footnotesize},
        ]
      \end{axis}


      \begin{groupplot}[
          group style={
            group size=1 by 3,
            vertical sep=0.5cm  
          },
          boxplot/draw direction=y,
          width=\mylocalwidth,
          height=\mylocalheight,
          ymin=1, ymax=5,
          xmin=0.5, xmax=8.5,
          xtick={1,...,8},
          xticklabels={},
          tick label style={font=\footnotesize},
          grid=major,
          boxplot/box extend=0.4,
          title style={yshift=-4pt, font=\footnotesize,scale=0.8},
          tick label style={font=\footnotesize, scale=0.8},
          label style={font=\footnotesize,scale=0.8},
          legend to name=devlegend,
          legend style={font=\footnotesize, legend columns=-1},
        ]

        \nextgroupplot[
          title={Helpfulness},
          ytick={1,3,5},
          yticklabels={1,3,5},
        ]
        \addplot+[
          RoyalBlue,
          line width=\myboxlinewidth,
          boxplot prepared={
            median=5.0,
            upper quartile=5.0,
            lower quartile=4.0,
            upper whisker=5.0,
            lower whisker=3.0,
          },
          solid,
        ] coordinates {};

        \addplot+[
          RoyalBlue,
          line width=\myboxlinewidth,
          boxplot prepared={
            median=5.0,
            upper quartile=5.0,
            lower quartile=5.0,
            upper whisker=5.0,
            lower whisker=5.0,
          },
          solid,
        ] coordinates {};
        \addplot[
          only marks,
          mark=*,
          mark size=1pt,
          draw=RoyalBlue,
          fill=RoyalBlue,
        ] coordinates {(2,3) (2,4)};

        \addplot+[
          RoyalBlue,
          line width=\myboxlinewidth,
          boxplot prepared={
            median=5.0,
            upper quartile=5.0,
            lower quartile=4.0,
            upper whisker=5.0,
            lower whisker=3.0,
          },
          solid,
        ] coordinates {};
        \addplot[
          only marks,
          mark=*,
          mark size=1pt,
          draw=RoyalBlue,
          fill=RoyalBlue,
        ] coordinates {(3,2)};

        \addplot+[
          RoyalBlue,
          line width=\myboxlinewidth,
          boxplot prepared={
            median=5.0,
            upper quartile=5.0,
            lower quartile=3.5,
            upper whisker=5.0,
            lower whisker=2.0,
          },
          solid,
        ] coordinates {};
        \addplot[
          only marks,
          mark=*,
          mark size=1pt,
          draw=RoyalBlue,
          fill=RoyalBlue,
        ] coordinates {(4,1)};

        \addplot+[
          RoyalBlue,
          line width=\myboxlinewidth,
          boxplot prepared={
            median=5.0,
            upper quartile=5.0,
            lower quartile=4.0,
            upper whisker=5.0,
            lower whisker=3.0,
          },
          solid,
        ] coordinates {};

        \addplot+[
          RoyalBlue,
          line width=\myboxlinewidth,
          boxplot prepared={
            median=4.0,
            upper quartile=5.0,
            lower quartile=4.0,
            upper whisker=5.0,
            lower whisker=2.0,
          },
          solid,
        ] coordinates {};

        \addplot+[
          RoyalBlue,
          line width=\myboxlinewidth,
          boxplot prepared={
            median=5.0,
            upper quartile=5.0,
            lower quartile=3.0,
            upper whisker=5.0,
            lower whisker=2.0,
          },
          solid,
        ] coordinates {};

        \addplot+[
          RoyalBlue,
          line width=\myboxlinewidth,
          boxplot prepared={
            median=4.0,
            upper quartile=5.0,
            lower quartile=4.0,
            upper whisker=5.0,
            lower whisker=3.0,
          },
          solid,
        ] coordinates {};
        \addplot[
          only marks,
          mark=*,
          mark size=1pt,
          draw=RoyalBlue,
          fill=RoyalBlue,
        ] coordinates {(8,2)};

        \addplot[semithick, red, sharp plot]
        coordinates {(1,4.45) (2,4.75) (3,4.2) (4,4.1) (5,4.5) (6,3.95) (7,3.95) (8,4.15)};
        \addlegendentry{Distribution}
        \addlegendentry{Mean}

        \nextgroupplot[
          title={Length},
          ylabel={Rating},
          ytick={1,3,5},
          yticklabels={1,3,5},
        ]
        \addplot+[
          RoyalBlue,
          line width=\myboxlinewidth,
          boxplot prepared={
            median=4.0,
            upper quartile=5.0,
            lower quartile=4.,
            upper whisker=5.0,
            lower whisker=3.0,
          },
          solid,
        ] coordinates {};

        \addplot+[
          RoyalBlue,
          line width=\myboxlinewidth,
          boxplot prepared={
            median=5.0,
            upper quartile=5.0,
            lower quartile=4.0,
            upper whisker=5.0,
            lower whisker=4.0,
          },
          solid,
        ] coordinates {};
        \addplot[
          only marks,
          mark=*,
          mark size=1pt,
          draw=RoyalBlue,
          fill=RoyalBlue,
        ] coordinates {(2,2)};

        \addplot+[
          RoyalBlue,
          line width=\myboxlinewidth,
          boxplot prepared={
            median=4.0,
            upper quartile=4.0,
            lower quartile=3.0,
            upper whisker=5.0,
            lower whisker=2.0,
          },
          solid,
        ] coordinates {};
        \addplot[
          only marks,
          mark=*,
          mark size=1pt,
          draw=RoyalBlue,
          fill=RoyalBlue,
        ] coordinates {(3,1)};

        \addplot+[
          RoyalBlue,
          line width=\myboxlinewidth,
          boxplot prepared={
            median=4.0,
            upper quartile=5.0,
            lower quartile=3.5,
            upper whisker=5.0,
            lower whisker=2.0,
          },
          solid,
        ] coordinates {};
        \addplot[
          only marks,
          mark=*,
          mark size=1pt,
          draw=RoyalBlue,
          fill=RoyalBlue,
        ] coordinates {(4,1)};

        \addplot+[
          RoyalBlue,
          line width=\myboxlinewidth,
          boxplot prepared={
            median=4.0,
            upper quartile=4.0,
            lower quartile=2.5,
            upper whisker=5.0,
            lower whisker=1.0,
          },
          solid,
        ] coordinates {};

        \addplot+[
          RoyalBlue,
          line width=\myboxlinewidth,
          boxplot prepared={
            median=2.5,
            upper quartile=3.5,
            lower quartile=2.0,
            upper whisker=4.0,
            lower whisker=1.0,
          },
          solid,
        ] coordinates {};

        \addplot+[
          RoyalBlue,
          line width=\myboxlinewidth,
          boxplot prepared={
            median=3.0,
            upper quartile=3.5,
            lower quartile=2.5,
            upper whisker=5.0,
            lower whisker=1.0,
          },
          solid,
        ] coordinates {};

        \addplot+[
          RoyalBlue,
          line width=\myboxlinewidth,
          boxplot prepared={
            median=3.0,
            upper quartile=3.5,
            lower quartile=2.0,
            upper whisker=5.0,
            lower whisker=1.0,
          },
          solid,
        ] coordinates {};
        \addplot[semithick, red, sharp plot]
        coordinates {(1,4.2) (2,4.55) (3,3.55) (4,4.05) (5,3.45) (6,2.7) (7,2.9) (8,3.0)};

        \nextgroupplot[
          title={Structure},
          ytick={1,3,5},
          yticklabels={1,3,5},
          xlabel={Question},
          xticklabels={1,2,3,4,5,6,7,8},
        ]
        \addplot+[
          RoyalBlue,
          line width=\myboxlinewidth,
          boxplot prepared={
            median=5.0,
            upper quartile=5.0,
            lower quartile=4.0,
            upper whisker=5.0,
            lower whisker=3.0,
          },
          solid,
        ] coordinates {};
        \addplot[
          only marks,
          mark=*,
          mark size=1pt,
          draw=RoyalBlue,
          fill=RoyalBlue,
        ] coordinates {(1,2)};

        \addplot+[
          RoyalBlue,
          line width=\myboxlinewidth,
          boxplot prepared={
            median=5.0,
            upper quartile=5.0,
            lower quartile=5.0,
            upper whisker=5.0,
            lower whisker=5.0,
          },
          solid,
        ] coordinates {};
        \addplot[
          only marks,
          mark=*,
          mark size=1pt,
          draw=RoyalBlue,
          fill=RoyalBlue,
        ] coordinates {(2,3) (2,4)};

        \addplot+[
          RoyalBlue,
          line width=\myboxlinewidth,
          boxplot prepared={
            median=5.0,
            upper quartile=5.0,
            lower quartile=4.0,
            upper whisker=5.0,
            lower whisker=3.0,
          },
          solid,
        ] coordinates {};
        \addplot[
          only marks,
          mark=*,
          mark size=1pt,
          draw=RoyalBlue,
          fill=RoyalBlue,
        ] coordinates {(3,2)};

        \addplot+[
          RoyalBlue,
          line width=\myboxlinewidth,
          boxplot prepared={
            median=5.0,
            upper quartile=5.0,
            lower quartile=4.0,
            upper whisker=5.0,
            lower whisker=3.0,
          },
          solid,
        ] coordinates {};
        \addplot[
          only marks,
          mark=*,
          mark size=1pt,
          draw=RoyalBlue,
          fill=RoyalBlue,
        ] coordinates {(4,1)};

        \addplot+[
          RoyalBlue,
          line width=\myboxlinewidth,
          boxplot prepared={
            median=5.0,
            upper quartile=5.0,
            lower quartile=4.0,
            upper whisker=5.0,
            lower whisker=3.0,
          },
          solid,
        ] coordinates {};
        \addplot[
          only marks,
          mark=*,
          mark size=1pt,
          draw=RoyalBlue,
          fill=RoyalBlue,
        ] coordinates {(5,2)};

        \addplot+[
          RoyalBlue,
          line width=\myboxlinewidth,
          boxplot prepared={
            median=4.0,
            upper quartile=4.0,
            lower quartile=3.0,
            upper whisker=5.0,
            lower whisker=2.0,
          },
          solid,
        ] coordinates {};

        \addplot+[
          RoyalBlue,
          line width=\myboxlinewidth,
          boxplot prepared={
            median=4.0,
            upper quartile=5.0,
            lower quartile=3.5,
            upper whisker=5.0,
            lower whisker=2.0,
          },
          solid,
        ] coordinates {};

        \addplot+[
          RoyalBlue,
          line width=\myboxlinewidth,
          boxplot prepared={
            median=4.0,
            upper quartile=4.0,
            lower quartile=3.0,
            upper whisker=5.0,
            lower whisker=2.0,
          },
          solid,
        ] coordinates {};
        \addplot[semithick, red, sharp plot]
        coordinates {(1,4.35) (2,4.8) (3,4.5) (4,4.4) (5,4.35) (6,3.65) (7,4.05) (8,3.75)};

      \end{groupplot}

    \end{tikzpicture}
  \end{subfigure}
  \begin{subfigure}{0.48\textwidth}
    \centering
    \begin{tikzpicture}

      \begin{axis}[
          hide axis,
          height=50pt,
          width=\mylocalwidth,
          title={\bfseries Developer},
          title style={yshift=26pt, font=\footnotesize},
        ]
      \end{axis}


      \begin{groupplot}[
          group style={
            group size=1 by 3,
            vertical sep=0.5cm  
          },
          boxplot/draw direction=y,
          width=\mylocalwidth,
          height=\mylocalheight,
          ymin=1, ymax=5,
          xmin=0.5, xmax=8.5,
          xtick={1,...,8},
          xticklabels={},
          tick label style={font=\footnotesize},
          grid=major,
          boxplot/box extend=0.4,
          title style={yshift=-4pt, font=\footnotesize,scale=0.8},
          tick label style={font=\footnotesize, scale=0.8},
          label style={font=\footnotesize,scale=0.8},
          legend to name=devlegend,
          legend style={font=\footnotesize, legend columns=-1},
        ]

        \nextgroupplot[
          title={Helpfulness},
          ytick={1,3,5},
          yticklabels={1,3,5},
        ]
        \addplot+[
          RoyalBlue,
          line width=\myboxlinewidth,
          boxplot prepared={
            median=4.0,
            upper quartile=5.0,
            lower quartile=3.5,
            upper whisker=5.0,
            lower whisker=2.0,
          },
          solid,
        ] coordinates {};

        \addplot+[
          RoyalBlue,
          line width=\myboxlinewidth,
          boxplot prepared={
            median=5.0,
            upper quartile=5.0,
            lower quartile=4.0,
            upper whisker=5.0,
            lower whisker=3.0,
          },
          solid,
        ] coordinates {};
        \addplot[
          only marks,
          mark=*,
          mark size=1pt,
          draw=RoyalBlue,
          fill=RoyalBlue,
        ] coordinates {(2,2)};

        \addplot+[
          RoyalBlue,
          line width=\myboxlinewidth,
          boxplot prepared={
            median=5.0,
            upper quartile=5.0,
            lower quartile=5.0,
            upper whisker=5.0,
            lower whisker=5.0,
          },
          solid,
        ] coordinates {};
        \addplot[
          only marks,
          mark=*,
          mark size=1pt,
          draw=RoyalBlue,
          fill=RoyalBlue,
        ] coordinates {(3,2) (3,4)};

        \addplot+[
          RoyalBlue,
          line width=\myboxlinewidth,
          boxplot prepared={
            median=5.0,
            upper quartile=5.0,
            lower quartile=4.0,
            upper whisker=5.0,
            lower whisker=3.0,
          },
          solid,
        ] coordinates {};
        \addplot[
          only marks,
          mark=*,
          mark size=1pt,
          draw=RoyalBlue,
          fill=RoyalBlue,
        ] coordinates {(4,2)};

        \addplot+[
          RoyalBlue,
          line width=\myboxlinewidth,
          boxplot prepared={
            median=5.0,
            upper quartile=5.0,
            lower quartile=4.0,
            upper whisker=5.0,
            lower whisker=3.0,
          },
          solid,
        ] coordinates {};
        \addplot[
          only marks,
          mark=*,
          mark size=1pt,
          draw=RoyalBlue,
          fill=RoyalBlue,
        ] coordinates {(5,2)};

        \addplot+[
          RoyalBlue,
          line width=\myboxlinewidth,
          boxplot prepared={
            median=5.0,
            upper quartile=5.0,
            lower quartile=4.0,
            upper whisker=5.0,
            lower whisker=3.0,
          },
          solid,
        ] coordinates {};
        \addplot[
          only marks,
          mark=*,
          mark size=1pt,
          draw=RoyalBlue,
          fill=RoyalBlue,
        ] coordinates {(6,2)};

        \addplot+[
          RoyalBlue,
          line width=\myboxlinewidth,
          boxplot prepared={
            median=5.0,
            upper quartile=5.0,
            lower quartile=4.0,
            upper whisker=5.0,
            lower whisker=3.0,
          },
          solid,
        ] coordinates {};
        \addplot[
          only marks,
          mark=*,
          mark size=1pt,
          draw=RoyalBlue,
          fill=RoyalBlue,
        ] coordinates {(7,2)};

        \addplot+[
          RoyalBlue,
          line width=\myboxlinewidth,
          boxplot prepared={
            median=4.0,
            upper quartile=5.0,
            lower quartile=4.0,
            upper whisker=5.0,
            lower whisker=3.0,
          },
          solid,
        ] coordinates {};

        \addplot[semithick, red, sharp plot]
        coordinates {(1,4.05) (2,4.55) (3,4.7) (4,4.4) (5,4.39) (6,4.4) (7,4.35) (8,4.25)};
        \addlegendentry{Distribution}
        \addlegendentry{Mean}

        \nextgroupplot[
          title={Length},
          ylabel={Rating},
          ytick={1,3,5},
          yticklabels={1,3,5},
        ]
        \addplot+[
          RoyalBlue,
          line width=\myboxlinewidth,
          boxplot prepared={
            median=4.0,
            upper quartile=5.0,
            lower quartile=3.5,
            upper whisker=5.0,
            lower whisker=2.0,
          },
          solid,
        ] coordinates {};
        \addplot[
          only marks,
          mark=*,
          mark size=1pt,
          draw=RoyalBlue,
          fill=RoyalBlue,
        ] coordinates {(1,1)};

        \addplot+[
          RoyalBlue,
          line width=\myboxlinewidth,
          boxplot prepared={
            median=4.0,
            upper quartile=5.0,
            lower quartile=3.0,
            upper whisker=5.0,
            lower whisker=2.0,
          },
          solid,
        ] coordinates {};

        \addplot+[
          RoyalBlue,
          line width=\myboxlinewidth,
          boxplot prepared={
            median=5.0,
            upper quartile=5.0,
            lower quartile=4.0,
            upper whisker=5.0,
            lower whisker=3.0,
          },
          solid,
        ] coordinates {};
        \addplot[
          only marks,
          mark=*,
          mark size=1pt,
          draw=RoyalBlue,
          fill=RoyalBlue,
        ] coordinates {(3,2)};

        \addplot+[
          RoyalBlue,
          line width=\myboxlinewidth,
          boxplot prepared={
            median=4.0,
            upper quartile=5.0,
            lower quartile=3.5,
            upper whisker=5.0,
            lower whisker=2.0,
          },
          solid,
        ] coordinates {};

        \addplot+[
          RoyalBlue,
          line width=\myboxlinewidth,
          boxplot prepared={
            median=4.0,
            upper quartile=5.0,
            lower quartile=4.0,
            upper whisker=5.0,
            lower whisker=3.0,
          },
          solid,
        ] coordinates {};

        \addplot+[
          RoyalBlue,
          line width=\myboxlinewidth,
          boxplot prepared={
            median=4.0,
            upper quartile=5.0,
            lower quartile=3.0,
            upper whisker=5.0,
            lower whisker=1.0,
          },
          solid,
        ] coordinates {};

        \addplot+[
          RoyalBlue,
          line width=\myboxlinewidth,
          boxplot prepared={
            median=3.0,
            upper quartile=4.5,
            lower quartile=2.5,
            upper whisker=5.0,
            lower whisker=1.0,
          },
          solid,
        ] coordinates {};

        \addplot+[
          RoyalBlue,
          line width=\myboxlinewidth,
          boxplot prepared={
            median=4.0,
            upper quartile=4.0,
            lower quartile=3.0,
            upper whisker=5.0,
            lower whisker=2.0,
          },
          solid,
        ] coordinates {};
        \addplot[
          only marks,
          mark=*,
          mark size=1pt,
          draw=RoyalBlue,
          fill=RoyalBlue,
        ] coordinates {(8,1)};

        \addplot[semithick, red, sharp plot]
        coordinates {(1,3.7) (2,3.85) (3,4.55) (4,4) (5,4.22) (6,3.85) (7,3.4) (8,3.55)};

        \nextgroupplot[
          title={Structure},
          ytick={1,3,5},
          yticklabels={1,3,5},
          xlabel={Question},
          xticklabels={1,2,3,4,5,6,7,8},
        ]
        \addplot+[
          RoyalBlue,
          line width=\myboxlinewidth,
          boxplot prepared={
            median=4.0,
            upper quartile=4.5,
            lower quartile=3.0,
            upper whisker=5.0,
            lower whisker=1.0,
          },
          solid,
        ] coordinates {};

        \addplot+[
          RoyalBlue,
          line width=\myboxlinewidth,
          boxplot prepared={
            median=5.0,
            upper quartile=5.0,
            lower quartile=4.0,
            upper whisker=5.0,
            lower whisker=3.0,
          },
          solid,
        ] coordinates {};
        \addplot[
          only marks,
          mark=*,
          mark size=1pt,
          draw=RoyalBlue,
          fill=RoyalBlue,
        ] coordinates {(2,2)};

        \addplot+[
          RoyalBlue,
          line width=\myboxlinewidth,
          boxplot prepared={
            median=5.0,
            upper quartile=5.0,
            lower quartile=4.0,
            upper whisker=5.0,
            lower whisker=4.0,
          },
          solid,
        ] coordinates {};
        \addplot[
          only marks,
          mark=*,
          mark size=1pt,
          draw=RoyalBlue,
          fill=RoyalBlue,
        ] coordinates {(3,2)};

        \addplot+[
          RoyalBlue,
          line width=\myboxlinewidth,
          boxplot prepared={
            median=5.0,
            upper quartile=5.0,
            lower quartile=4.0,
            upper whisker=5.0,
            lower whisker=3.0,
          },
          solid,
        ] coordinates {};
        \addplot[
          only marks,
          mark=*,
          mark size=1pt,
          draw=RoyalBlue,
          fill=RoyalBlue,
        ] coordinates {(4,2)};

        \addplot+[
          RoyalBlue,
          line width=\myboxlinewidth,
          boxplot prepared={
            median=5.0,
            upper quartile=5.0,
            lower quartile=4.0,
            upper whisker=5.0,
            lower whisker=3.0,
          },
          solid,
        ] coordinates {};

        \addplot+[
          RoyalBlue,
          line width=\myboxlinewidth,
          boxplot prepared={
            median=4.0,
            upper quartile=5.0,
            lower quartile=3.0,
            upper whisker=5.0,
            lower whisker=2.0,
          },
          solid,
        ] coordinates {};

        \addplot+[
          RoyalBlue,
          line width=\myboxlinewidth,
          boxplot prepared={
            median=3.8,
            upper quartile=5.0,
            lower quartile=3.5,
            upper whisker=5.0,
            lower whisker=3.0,
          },
          solid,
        ] coordinates {};

        \addplot+[
          RoyalBlue,
          line width=\myboxlinewidth,
          boxplot prepared={
            median=4.0,
            upper quartile=5.0,
            lower quartile=4.0,
            upper whisker=5.0,
            lower whisker=3.0,
          },
          solid,
        ] coordinates {};
        \addplot[
          only marks,
          mark=*,
          mark size=1pt,
          draw=RoyalBlue,
          fill=RoyalBlue,
        ] coordinates {(8,2)};

        \addplot[semithick, red, sharp plot]
        coordinates {(1,3.8) (2,4.3) (3,4.4) (4,4.3) (5,4.61) (6,3.8) (7,4.1) (8,4.2)};

      \end{groupplot}

    \end{tikzpicture}
  \end{subfigure}


  \pgfplotslegendfromname{devlegend}
  \caption{Subjective evaluation of answers for user groups worker and developer (boxplots: ranking distribution per question, red line: mean, blue dot: outlier)}
  \label{fig:work-dev-boxplots}
\end{figure*}

\label{subsection:eval_methodology}
We evaluated our approach using two complementary approaches: a user-based study and a system-oriented evaluation. In the user-based study, 20 participants with professional experience in AI assessed the quality of system outputs. Each participant rated two potential user roles (\textit{developer} and \textit{worker}), and eight representative answers per role. The questions are ordered by increasing answer complexity, and each answer is rated according to the criteria \textit{helpfulness and understandability}, \textit{structure}, and \textit{length appropriateness} on a five-point Likert scale (1 = lowest, 5 = highest). All prompts and system responses were fixed across participants, and the factual correctness of the answers was verified beforehand so that ratings targeted textual presentation only. The results are analyzed using descriptive statistics and Kendall’s~$\tau$ coefficient to examine internal consistency.

In addition, we conducted a structured evaluation of our approach, using a catalog of 24 questions (available in the GitHub-Repository, cf. fn~\ref{fn:github_repo}, p.~\pageref{fn:github_repo})) targeting aspects beyond the primary use case. The questions cover seven categories: ambiguity, contradictions, out-of-scope queries, overgeneralization and bias, instructional confusion, complex cross-referencing, and prompt-injection attempts. For each item, an expected response pattern was defined in advance and used to qualitatively assess robustness and failure modes of the prototype.

\subsection{User-Based Evaluation}\label{subsection:user_based_evaluation}

The user based evaluation in Fig.~\ref{fig:work-dev-boxplots} reveals noticeable variations across questions and criteria, with clear patterns emerging between user groups. \textit{Helpfulness and understandability} receive stable mid-to-high scores. \textit{Developer} ratings exhibit slightly higher means and a reduced spread, indicating more homogeneous judgments, whereas \textit{worker} ratings show broader distributions and occasional outliers, suggesting more diverse perceptions. \textit{Length appropriateness} displays the strongest fluctuations. \textit{Worker} ratings decrease for more complex questions, while \textit{developer} ratings remain more stable but still vary between questions, indicating sensitivity to differences in verbosity or conciseness. \textit{Structure} ratings follow a similar trend, with overall mid-range values. \textit{Developer} evaluations again show tighter distributions and slightly higher central tendencies, whereas \textit{worker} ratings contain more outliers and wider interquartile ranges.

Across all criteria, no extreme deviations or systematic failures appear. The mean curves remain within a narrow band, showing that the system maintains a consistent level of perceived quality, although outliers in both groups indicate that certain answers diverge from expectations, particularly for the \textit{worker} role.

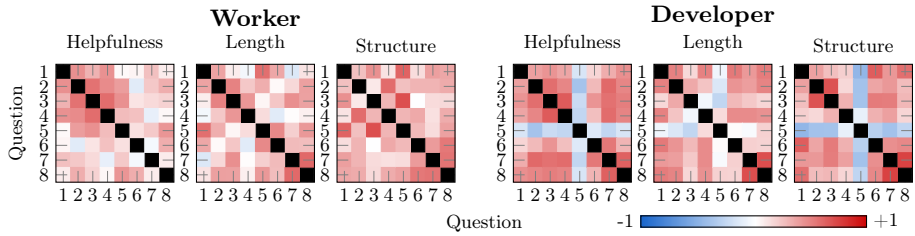
\begin{figure*}[t]
\footnotesize
\centering

\def\heatplotsize{3.15cm}
\newcommand{\NaNcoords}{%
  (0,0) (1,1) (2,2) (3,3) (4,4) (5,5) (6,6) (7,7)%
}

\newcommand{\NaNOverlay}{%
  \addplot[
    only marks,
    mark=square*,
    mark size=2.5,
    draw=black,
    fill=black,
  ]
  coordinates {\NaNcoords};
}
\newcommand{\WorkerHelpHeatmap}{
  \addplot[matrix plot, point meta=explicit]
  coordinates {
    (0,0) [nan]      (1,0) [0.4367]  (2,0) [0.3066]  (3,0) [0.4325]
    (4,0) [0.0272]   (5,0) [0.0482]  (6,0) [0.2492]  (7,0) [0.0763]

    (0,1) [0.4367]   (1,1) [nan]     (2,1) [0.5530]  (3,1) [0.4564]
    (4,1) [0.3879]   (5,1) [-0.1253] (6,1) [0.2192]  (7,1) [0.3195]

    (0,2) [0.3066]   (1,2) [0.5530]  (2,2) [nan]     (3,2) [0.5723]
    (4,2) [0.4356]   (5,2) [0.1083]  (6,2) [0.1470]  (7,2) [0.1796]

    (0,3) [0.4325]   (1,3) [0.4564]  (2,3) [0.5723]  (3,3) [nan]
    (4,3) [0.2673]   (5,3) [0.2455]  (6,3) [-0.0982] (7,3) [0.0345]

    (0,4) [0.0272]   (1,4) [0.3879]  (2,4) [0.4356]  (3,4) [0.2673]
    (4,4) [nan]      (5,4) [0.1233]  (6,4) [0.2137]  (7,4) [0.4165]

    (0,5) [0.0482]   (1,5) [-0.1253] (2,5) [0.1083]  (3,5) [0.2455]
    (4,5) [0.1233]   (5,5) [nan]     (6,5) [-0.0438] (7,5) [0.3235]

    (0,6) [0.2492]   (1,6) [0.2192]  (2,6) [0.1470]  (3,6) [-0.0982]
    (4,6) [0.2137]   (5,6) [-0.0438] (6,6) [nan]     (7,6) [0.0770]

    (0,7) [0.0763]   (1,7) [0.3195]  (2,7) [0.1796]  (3,7) [0.0345]
    (4,7) [0.4165]   (5,7) [0.3235]  (6,7) [0.0770]  (7,7) [nan]
  };
  \NaNOverlay
}
\newcommand{\WorkerLengthHeatmap}{
  \addplot[matrix plot, point meta=explicit]
  coordinates {
    (0,0) [nan]      (1,0) [0.3615]   (2,0) [0.0737]   (3,0) [-0.0501]
    (4,0) [0.5673]   (5,0) [0.3491]   (6,0) [-0.1491]  (7,0) [0.1127]

    (0,1) [0.3615]   (1,1) [nan]      (2,1) [0.4927]   (3,1) [0.2201]
    (4,1) [0.3210]   (5,1) [0.0263]   (6,1) [0.1573]   (7,1) [0.3736]

    (0,2) [0.0737]   (1,2) [0.4927]   (2,2) [nan]      (3,2) [0.4134]
    (4,2) [0.0436]   (5,2) [0.2280]   (6,2) [0.4396]   (7,2) [0.3702]

    (0,3) [-0.0501]  (1,3) [0.2201]   (2,3) [0.4134]   (3,3) [nan]
    (4,3) [0.1408]   (5,3) [-0.1273]  (6,3) [0.0149]   (7,3) [0.0653]

    (0,4) [0.5673]   (1,4) [0.3210]   (2,4) [0.0436]   (3,4) [0.1408]
    (4,4) [nan]      (5,4) [0.4650]   (6,4) [0.1912]   (7,4) [0.4072]

    (0,5) [0.3491]   (1,5) [0.0263]   (2,5) [0.2280]   (3,5) [-0.1273]
    (4,5) [0.4650]   (5,5) [nan]      (6,5) [0.2677]   (7,5) [0.2601]

    (0,6) [-0.1491]  (1,6) [0.1573]   (2,6) [0.4396]   (3,6) [0.0149]
    (4,6) [0.1912]   (5,6) [0.2677]   (6,6) [nan]      (7,6) [0.5902]

    (0,7) [0.1127]   (1,7) [0.3736]   (2,7) [0.3702]   (3,7) [0.0653]
    (4,7) [0.4072]   (5,7) [0.2601]   (6,7) [0.5902]   (7,7) [nan]
  };
  \NaNOverlay
}
\newcommand{\WorkerStructureHeatmap}{
  \addplot[matrix plot, point meta=explicit]
  coordinates {
    (0,0) [nan]      (1,0) [0.2518]   (2,0) [0.4357]   (3,0) [0.1129]
    (4,0) [0.6134]   (5,0) [0.1351]   (6,0) [0.3787]   (7,0) [0.3471]

    (0,1) [0.2518]   (1,1) [nan]      (2,1) [0.0952]   (3,1) [0.5778]
    (4,1) [0.2518]   (5,1) [0.2978]   (6,1) [0.2956]   (7,1) [0.3310]

    (0,2) [0.4357]   (1,2) [0.0952]   (2,2) [nan]      (3,2) [0.2946]
    (4,2) [0.6898]   (5,2) [0.0000]   (6,2) [0.1449]   (7,2) [0.1534]

    (0,3) [0.1129]   (1,3) [0.5778]   (2,3) [0.2946]   (3,3) [nan]
    (4,3) [0.2916]   (5,3) [0.3381]   (6,3) [0.1236]   (7,3) [0.3267]

    (0,4) [0.6134]   (1,4) [0.2518]   (2,4) [0.6898]   (3,4) [0.2916]
    (4,4) [nan]      (5,4) [0.2782]   (6,4) [0.1420]   (7,4) [0.3708]

    (0,5) [0.1351]   (1,5) [0.2978]   (2,5) [0.0000]   (3,5) [0.3381]
    (4,5) [0.2782]   (5,5) [nan]      (6,5) [0.3955]   (7,5) [0.5299]

    (0,6) [0.3787]   (1,6) [0.2956]   (2,6) [0.1449]   (3,6) [0.1236]
    (4,6) [0.1420]   (5,6) [0.3955]   (6,6) [nan]      (7,6) [0.5481]

    (0,7) [0.3471]   (1,7) [0.3310]   (2,7) [0.1534]   (3,7) [0.3267]
    (4,7) [0.3708]   (5,7) [0.5299]   (6,7) [0.5481]   (7,7) [nan]
  };
  \NaNOverlay
}
\newcommand{\DevHelpHeatmap}{
  \addplot[matrix plot, point meta=explicit]
  coordinates {
    (0,0) [nan]      (1,0) [0.3748]   (2,0) [0.4206]   (3,0) [0.3562]
    (4,0) [-0.1673]  (5,0) [0.1693]   (6,0) [0.3235]   (7,0) [0.4423]

    (0,1) [0.3748]   (1,1) [nan]      (2,1) [0.5701]   (3,1) [0.4780]
    (4,1) [-0.3917]  (5,1) [0.2719]   (6,1) [0.5894]   (7,1) [0.4768]

    (0,2) [0.4206]   (1,2) [0.5701]   (2,2) [nan]      (3,2) [0.6436]
    (4,2) [-0.2295]  (5,2) [0.4234]   (6,2) [0.5600]   (7,2) [0.5067]

    (0,3) [0.3562]   (1,3) [0.4780]   (2,3) [0.6436]   (3,3) [nan]
    (4,3) [-0.2632]  (5,3) [0.2555]   (6,3) [0.5778]   (7,3) [0.4835]

    (0,4) [-0.1673]  (1,4) [-0.3917]  (2,4) [-0.2295]  (3,4) [-0.2632]
    (4,4) [nan]      (5,4) [-0.1398]  (6,4) [-0.2632]  (7,4) [-0.1651]

    (0,5) [0.1693]   (1,5) [0.2719]   (2,5) [0.4234]   (3,5) [0.2555]
    (4,5) [-0.1398]  (5,5) [nan]      (6,5) [0.5323]   (7,5) [0.2956]

    (0,6) [0.3235]   (1,6) [0.5894]   (2,6) [0.5600]   (3,6) [0.5778]
    (4,6) [-0.2632]  (5,6) [0.5323]   (6,6) [nan]      (7,6) [0.6199]

    (0,7) [0.4423]   (1,7) [0.4768]   (2,7) [0.5067]   (3,7) [0.4835]
    (4,7) [-0.1651]  (5,7) [0.2956]   (6,7) [0.6199]   (7,7) [nan]
  };
  \NaNOverlay
}
\newcommand{\DevLengthHeatmap}{%
  \addplot[matrix plot, point meta=explicit]
  coordinates {
    (0,0) [nan]      (1,0) [0.4952]   (2,0) [0.1570]   (3,0) [0.4944]
    (4,0) [-0.0619]  (5,0) [0.5271]   (6,0) [0.4033]   (7,0) [0.5140]

    (0,1) [0.4952]   (1,1) [nan]      (2,1) [0.3406]   (3,1) [0.2337]
    (4,1) [-0.1549]  (5,1) [0.4211]   (6,1) [0.3813]   (7,1) [0.4859]

    (0,2) [0.1570]   (1,2) [0.3406]   (2,2) [nan]      (3,2) [-0.0539]
    (4,2) [-0.2664]  (5,2) [0.2765]   (6,2) [0.2180]   (7,2) [0.1648]

    (0,3) [0.4944]   (1,3) [0.2337]   (2,3) [-0.0539]  (3,3) [nan]
    (4,3) [0.1431]   (5,3) [0.4552]   (6,3) [0.4557]   (7,3) [0.4351]

    (0,4) [-0.0619]  (1,4) [-0.1549]  (2,4) [-0.2664]  (3,4) [0.1431]
    (4,4) [nan]      (5,4) [0.0197]   (6,4) [-0.0282]  (7,4) [0.1831]

    (0,5) [0.5271]   (1,5) [0.4211]   (2,5) [0.2765]   (3,5) [0.4552]
    (4,5) [0.0197]   (5,5) [nan]      (6,5) [0.1886]   (7,5) [0.1742]

    (0,6) [0.4033]   (1,6) [0.3813]   (2,6) [0.2180]   (3,6) [0.4557]
    (4,6) [-0.0282]  (5,6) [0.1886]   (6,6) [nan]      (7,6) [0.6963]

    (0,7) [0.5140]   (1,7) [0.4859]   (2,7) [0.1648]   (3,7) [0.4351]
    (4,7) [0.1831]   (5,7) [0.1742]   (6,7) [0.6963]   (7,7) [nan]
  };
  \NaNOverlay
}
\newcommand{\DevStructureHeatmap}{
  \addplot[matrix plot, point meta=explicit]
  coordinates {
    (0,0) [nan]      (1,0) [0.3728]   (2,0) [0.2549]   (3,0) [0.1904]
    (4,0) [-0.4738]  (5,0) [0.6386]   (6,0) [0.4106]   (7,0) [0.5655]

    (0,1) [0.3728]   (1,1) [nan]      (2,1) [0.6790]   (3,1) [0.1552]
    (4,1) [-0.3989]  (5,1) [0.3547]   (6,1) [0.4056]   (7,1) [0.2657]

    (0,2) [0.2549]   (1,2) [0.6790]   (2,2) [nan]      (3,2) [0.3038]
    (4,2) [-0.3921]  (5,2) [0.5194]   (6,2) [0.5128]   (7,2) [0.2840]

    (0,3) [0.1904]   (1,3) [0.1552]   (2,3) [0.3038]   (3,3) [nan]
    (4,3) [-0.0956]  (5,3) [0.3084]   (6,3) [0.3732]   (7,3) [0.1661]

    (0,4) [-0.4738]  (1,4) [-0.3989]  (2,4) [-0.3921]  (3,4) [-0.0956]
    (4,4) [nan]      (5,4) [-0.3068]  (6,4) [-0.3966]  (7,4) [-0.2695]

    (0,5) [0.6386]   (1,5) [0.3547]   (2,5) [0.5194]   (3,5) [0.3084]
    (4,5) [-0.3068]  (5,5) [nan]      (6,5) [0.4426]   (7,5) [0.4680]

    (0,6) [0.4106]   (1,6) [0.4056]   (2,6) [0.5128]   (3,6) [0.3732]
    (4,6) [-0.3966]  (5,6) [0.4426]   (6,6) [nan]      (7,6) [0.7424]

    (0,7) [0.5655]   (1,7) [0.2657]   (2,7) [0.2840]   (3,7) [0.1661]
    (4,7) [-0.2695]  (5,7) [0.4680]   (6,7) [0.7424]   (7,7) [nan]
  };
  \NaNOverlay
}

\begin{tikzpicture}
  \begin{groupplot}[
      group style={
        group name=heatmaps,
        group size=7 by 1,
        horizontal sep=0.3cm
      },
      width=\heatplotsize,
      height=\heatplotsize,
      axis equal image,
      axis on top,
      enlargelimits=false,
      colormap name=bluewhiteRed,
      point meta min=-1,
      point meta max=1,
      ticks=both,
      xtick={0,...,7},
      xticklabels={1,2,3,4,5,6,7,8},
      ytick={0,...,7},
      yticklabels={1,2,3,4,5,6,7,8},
      y dir=reverse,
      title style={yshift=-4pt, font=\footnotesize, scale=0.8},
      label style={font=\footnotesize, scale=0.8},
      tick label style={font=\footnotesize, scale=0.8},
    ]

    \nextgroupplot[
      title={Helpfulness},
      ylabel={Question},
    ]
    \WorkerHelpHeatmap

    \nextgroupplot[
      title={Length},
    ]
    \WorkerLengthHeatmap

    \nextgroupplot[
      title={Structure},
    ]
    \WorkerStructureHeatmap
    \nextgroupplot[
      hide axis,
      width=50pt,
      height=\heatplotsize,
      enlargelimits=false,
    ]

    \nextgroupplot[
      title={Helpfulness},
    ]
    \DevHelpHeatmap

    \nextgroupplot[
      title={Length},
    ]
    \DevLengthHeatmap

    \nextgroupplot[
      title={Structure},
    ]
    \DevStructureHeatmap

  \end{groupplot}

  \coordinate (heatmapRight) at (heatmaps c7r1.south east);

  \node[
    anchor=south,
    yshift=1.2em,
    font=\footnotesize\bfseries
  ] at (heatmaps c2r1.north)
  {Worker};

  \node[
    anchor=south,
    yshift=1.2em,
    font=\footnotesize\bfseries
  ] at (heatmaps c6r1.north)
  {Developer};

  \node[anchor=north, yshift=-3.2em, font=\footnotesize, scale=0.8]   
  at ($(heatmaps c4r1.south)$)
  {Question};

  \node[
    anchor=north west,
    yshift=-0.1em,
    xshift=0.4cm
  ] at (heatmaps c5r1.south)
  {
    \begin{tikzpicture}[baseline={(current bounding box.north)}]

      \begin{axis}[
          name=cbaxis,
          hide axis,
          width=3cm,
          height=0.15cm,
          scale only axis,
          xmin=-1, xmax=1,
          ymin=0, ymax=1,
          colormap name=bluewhiteRed,
          point meta min=-1,
          point meta max=1,
          colorbar horizontal,
          colorbar style={
            width=3cm,
            height=0.15cm,
            xtick=\empty,
          },
        ]
        \addplot[draw=none] coordinates {(-1,0) (1,0)};
      \end{axis}

      \node[
        anchor=east,
        font=\footnotesize,
        scale=0.8,
        yshift=-3.4ex
      ] at (cbaxis.outer west) {-1};

      \node[
        anchor=west,
        font=\footnotesize,
        scale=0.8,
        yshift=-3.4ex
      ] at (cbaxis.outer east) {+1};

    \end{tikzpicture}
  };
\end{tikzpicture}
\caption{Kendall's $\tau$ correlation matrices for ratings of user roles. Shows pairwise correlations; black diagonal cells indicate undefined self-correlations (NaN)}\label{fig:work-dev-kandeltau}

\end{figure*}

To assess interpretation consistency, Kendall’s $\tau$ was computed for all items in both groups, as shown in Fig.~\ref{fig:work-dev-kandeltau}. For the \textit{developer} role, most questions exhibit positive correlations in \textit{helpfulness}, \textit{understandability}, and \textit{structure}, indicating a shared interpretation. Question~5 is the exception, with weak or negative correlations, suggesting that it was interpreted differently. The \textit{worker} group follows a similar but weaker pattern. \textit{Helpfulness} and \textit{understandability} remain largely aligned, with Question~6 displaying the least stability. \textit{Length appropriateness} is more heterogeneous, as the \textit{worker} role shows greater individual variation. \textit{Structure} retains strong positive correlations, indicating stable agreement.

Overall, the correlation patterns in Fig.~\ref{fig:work-dev-kandeltau} indicate that both user roles use the scales in a generally consistent way. Only two questions stand out: Question~5 for \textit{developer} and Question~6 for \textit{worker}, which suggests that these items may need refinement. These correlations support the observations from the ratings in Fig.~\ref{fig:work-dev-boxplots}: the scales are used in a broadly consistent way, and deviations in individual questions match the variations seen in the subjective scores.

\subsection{Evaluation of Robustness and Stress Tests}

A structured evaluation of the prototype was conducted to assess robustness beyond the primary use case, focusing on typical failure modes observed in RAG-based systems. All question–answer pairs and the complete evaluation are provided in the GitHub repository (see fn.~\ref{fn:github_repo}, p.~\pageref{fn:github_repo}). Each example contains the full question, the generated \llm \ answer, the expected characteristics of a suitable response, and a short section summarizing the evaluation of the interaction. The following section presents the observed behavior across the tested categories.

The analysis reveals strengths and limitations in system responses. Ambiguous queries are commonly resolved through implicit assumption rather than explicit clarification, yielding technically correct but prematurely optimized responses (see Fig.~\ref{lst:example_ambiguous}). Contradictions and false premises are largely identified and corrected using ontology-grounded evidence, with only occasional failures when inconsistencies are implicitly encoded in the knowledge structure. Out-of-scope queries are reliably rejected when they clearly involve external domains; however, domain and task limitations are not consistently communicated for broader or abstract requests, leading to overextended responses. Absolute or biased framings are sometimes challenged through task-specific metrics and trade-off discussions, yet are not rejected systematically, which can unintentionally reinforce biased premises. A weakness lies in capability awareness: the system frequently overestimates its own executability and provides guidance for actions it cannot perform, indicating insufficient enforcement of capability boundaries.

\begin{figure}[H]
\begin{lstlisting}[style=promptstyle]
#Question
What task is easier?

#Expected
Clarify what 'easier' means (e.g., for the robot, for modeling); possibly mention models and compare.

#Answer
Based on the ontology information provided, the task ScrewPicking appears to be easier compared to ScrewPlacement. Here's why:

ScrewPicking:

The task involves picking up screws from different positions.
It does not mention additional complexities like angles or precise
placement.
There are no connected models or algorithms mentioned, which might
indicate that it is a simpler task that does not require advanced modeling.

ScrewPlacement:

The task involves placing screws into holes at different angles, which introduces additional complexity in terms of precision, orientation, and alignment.
Multiple models (logistic regression, decision tree, random forest
and neural network) are connected to this task, suggesting that it
requires more sophisticated approaches to achieve.

In summary, ScrewPicking is likely the easier task as it involves fewer variables and complexities compared to ScrewPlacement, which
requires precise handling and placement at varying angles.

#Notes
- No clarification on whats 'easier'
- Reasonable assumption based on task complexity, compares both
  tasks
- Did not mention model performances or learning difficulty
- Assumes ontology structure reflects task difficulty

\end{lstlisting}
\caption{Example: Ambiguous or Underspecified Request}\label{lst:example_ambiguous}
\end{figure}

In contrast, complex cross-referencing across tasks, models, and datasets is handled in a structured and internally consistent manner when the required metadata is available, with errors arising from unsupported fine-grained assumptions. Prompt-injection attempts further expose vulnerabilities, as adversarial instructions are not consistently rejected and may partially succeed before factual grounding is restored.

Overall, the evaluation indicates strong factual grounding and retrieval accuracy for well-scoped queries, while systematic weaknesses persist in interaction discipline, scope enforcement, capability modeling, and adversarial framing resistance, reflecting characteristic RAG failure modes rather than isolated errors.

\section{Discussion}

The evaluation shows that the proposed approach produces explanations that are perceived as clear, well structured, and useful across user roles. Ratings remain stable at mid-to-high levels. \textit{Developer} assessments exhibit higher homogeneity, whereas \textit{Worker} ratings show variability, indicating role-dependent expectations. Length appropriateness displays the highest dispersion, suggesting sensitivity to question complexity and user background. Kendall’s $\tau$ confirms strong internal consistency of ratings in both groups, with only isolated deviations.

Higher and more consistent ratings in the \textit{developer} role can be attributed to greater familiarity with technical terminology and conceptual framing. In contrast, feedback from \textit{worker} participants highlights occasional confusion when domain-specific terms are not explicitly explained. This suggests that explanation quality could be improved by role-aware adaptation of technical details.

Participants further expressed a preference for concise summaries placed at the beginning of responses, followed by optional elaboration. The current tendency toward detailed explanations was perceived as occasionally unnecessary and, in some cases, counterproductive. This indicates the need for improved control over information density and response length.

Despite these limitations, our approach demonstrates robust performance under targeted stress tests, maintaining reliable factual grounding and coherent cross-referencing across tasks, models, and datasets. The identified weaknesses correspond to commonly reported limitations of RAG-based systems.

\section{Summary}
Explaining ML results in a user-friendly manner remains challenging in XAI, particularly in manufacturing settings where domain context is essential for trust and decision-making. This paper presents a method that enhances interpretability by combining a KG with an LLM in a Graph-RAG workflow, storing domain knowledge alongside ML results and their corresponding explanations. Relevant triplets are selectively retrieved from the KG and provided as structured context to the LLM to generate user-friendly explanations grounded in domain semantics. We evaluate the method in a manufacturing use case using the XAI Question Bank, complemented by a user-based study and a structured system evaluation to assess both perceived quality and robustness. The results indicate that the generated explanations are perceived as clear, well structured, and useful across user roles, while also revealing typical RAG-related limitations in scope enforcement and interaction discipline. Overall, the findings suggest that KG-enhanced LLMs during inference provide a feasible pathway toward more actionable and practically usable post-hoc explainability in manufacturing.

Future work should focus on improved prompt design, explicit length constraints, and adaptive explanation strategies, as well as stronger alignment mechanisms and automated KG construction and maintenance. In addition, fine-tuning small open-source \llm s for effective querying of the KG, potentially replacing \texttt{GPT}-based components, represents a promising direction.

\begin{credits}
\subsubsection{Funding.}
Parts of the described research were funded by the Carl Zeiss Foundation under the auspices of the project KI-basierter digitaler Zwilling (KIDZ).
\subsubsection{\discintname}
The authors declare no conflict of interest.
\end{credits}

\bibliographystyle{splncs04}
\bibliography{bibliography}

@article{8466590,
  title = {Peeking inside the Black-Box: A Survey on Explainable Artificial Intelligence ({{XAI}})},
  author = {Adadi, Amina and Berrada, Mohammed},
  year = {2018},
  journal = {IEEE access : practical innovations, open solutions},
  volume = {6},
  pages = {52138--52160}
}

@article{Pan2024,
  title = {Unifying Large Language Models and Knowledge Graphs: {{A}} Roadmap},
  author = {Pan, Shirui and Luo, Linhao and Wang, Yufei and Chen, Chen and Wang, Jiapu and Wu, Xindong},
  year = {2024},
  journal = {IEEE Transactions on Knowledge and Data Engineering},
  volume = {36},
  number = {7},
  pages = {3580--3599}
}

@article{rony2022sgpt,
  title = {{{SGPT}}: A Generative Approach for {{SPARQL}} Query Generation from Natural Language Questions},
  author = {Rony, Md Rashad Al Hasan and Kumar, Uttam and Teucher, Roman and Kovriguina, Liubov and Lehmann, Jens},
  year = {2022},
  journal = {IEEE access : practical innovations, open solutions},
  volume = {10},
  pages = {70712--70723}
}

@article{wilson2023integration,
  title = {Integration of {{ML-schema}} with Machine Learning Platforms},
  author = {Wilson, J. and Liu, H.},
  year = {2023},
  journal = {Journal of Artificial Intelligence Research},
  volume = {10},
  number = {1},
  pages = {40--60}
}

@article{9429985,
  title = {Informed Machine Learning -- a Taxonomy and Survey of Integrating Prior Knowledge into Learning Systems},
  author = {{von Rueden}, Laura and Mayer, Sebastian and Beckh, Katharina and Georgiev, Bogdan and Giesselbach, Sven and Heese, Raoul and Kirsch, Birgit and Pfrommer, Julius and Pick, Annika and Ramamurthy, Rajkumar and Walczak, Michal and Garcke, Jochen and Bauckhage, Christian and Schuecker, Jannis},
  year = {2023},
  journal = {IEEE Transactions on Knowledge and Data Engineering},
  volume = {35},
  number = {1},
  pages = {614--633}
}

@article{BARREDOARRIETA202082,
  title = {Explainable {{Artificial Intelligence}} ({{XAI}}): {{Concepts}}, Taxonomies, Opportunities and Challenges toward Responsible {{AI}}},
  author = {Arrieta, Barredo Alejandro and {D{\'i}az-Rodr{\'i}guez}, Natalia and Del Ser, Javier and Bennetot, Adrien and Tabik, Siham and Barbado, Alberto and Garcia, Salvador and {Gil-Lopez}, Sergio and Molina, Daniel and Benjamins, Richard and Chatila, Raja and Herrera, Francisco},
  year = {2020},
  journal = {Information Fusion},
  volume = {58},
  pages = {82--115}
}

@article{BENHANIFIA2025200501,
  title = {Systematic Review of Predictive Maintenance Practices in the Manufacturing Sector},
  author = {Benhanifia, Abdeldjalil and Cheikh, Zied Ben and Oliveira, Paulo Moura and Valente, Antonio and Lima, Jos{\'e}},
  year = {2025},
  journal = {Intelligent Systems with Applications},
  volume = {26},
  pages = {200501}
}

@article{garcia2023ontologies,
  title = {The Role of Ontologies in Data Mining and Machine Learning},
  author = {Garcia, L. and Sanchez, M.},
  year = {2023},
  journal = {International Journal of Machine Learning},
  volume = {12},
  number = {4},
  pages = {200--220}
}

@article{hoepken2025KIDZ,
  author = {Wolfram Höpken and Ralf Stetter and Markus Pfeil and Thomas Bayer and Bernd Michelberger and Timo Schuchter and Alexander Lohr},
  title = {Digital Twins Using Semantic Modeling and AI},
  journal = {Industry 4.0 Science},
  volume = {41},
  number = {2},
  year = {2023},
  pages = {30-36}
}

@article{khan2023evaluating,
  title = {Evaluating {{ML-schema}}: {{An}} Empirical Study on Data Mining Interoperability},
  author = {Khan, F. and Lee, C.},
  year = {2023},
  journal = {Journal of Data Science and Technology},
  volume = {8},
  number = {2},
  pages = {50--70}
}

@article{kumar2019ontologies,
  title = {Ontologies for Industry 4.0},
  author = {Sampath Kumar, Veera Ragavan and Khamis, Alaa and Fiorini, Sandro and Carbonera, Joel Lu{\'i}s and Alarcos, Alberto Olivares and Habib, Maki and Goncalves, Paulo and Li, Howard and Olszewska, Joanna Isabelle},
  year = {2019},
  journal = {The Knowledge Engineering Review},
  volume = {34},
  pages = {e17}
}

@article{Lan2024LLM4QA,
  title = {{{LLM4QA}}: {{Leveraging}} Large Language Model for Efficient Knowledge Graph Reasoning with {{SPARQL}} Query},
  author = {Lan, Mingjing and Xia, Yi and Zhou, Gang and Huang, Ningbo and Li, Zhufeng and Wu, Hao},
  year = {2024},
  journal = {Journal of Advances in Information Technology},
  volume = {15},
  number = {10},
  pages = {1157--1162}
}

@article{liang2021querying,
  title = {Querying Knowledge Graphs in Natural Language},
  author = {Liang, Shiqi and Stockinger, Kurt and {Mendes de Farias}, Tarcisio and Anisimova, Maria and Gil, Manuel},
  year = {2021},
  journal = {Journal of Big Data},
  volume = {8},
  number = {3},
  pages = {3}
}

@article{mlschema:201610,
  title = {{{ML-Schema}}: {{Exposing}} the Semantics of Machine Learning with Schemas and Ontologies},
  author = {Publio, Gustavo Correa and Esteves, Diego and {\L}awrynowicz, Agnieszka and Panov, Pan{\v c}e and Soldatova, Larisa N. and Soru, Tommaso and Vanschoren, Joaquin and Zafar, Hamid},
  year = {2018},
  journal = {CoRR},
  volume = {abs/1807.05351},
  archiveprefix = {arXiv}
}

@inproceedings{dasoulas2024mlsea,
  title = {{{MLSea}}: A Semantic Layer for Discoverable Machine Learning},
  booktitle = {The Semantic Web -- {{ESWC}} 2024},
  author = {Dasoulas, Ioannis and others},
  year = {2024},
  series = {Lecture Notes in Computer Science},
  publisher = {Springer}
}

@inproceedings{ijcai2018p777,
  title = {Ontology-Based Data Access: A Survey},
  booktitle = {Proceedings of the Twenty-Seventh International Joint Conference on Artificial Intelligence, {{IJCAI-18}}},
  author = {Xiao, Guohui and Calvanese, Diego and Kontchakov, Roman and Lembo, Domenico and Poggi, Antonella and Rosati, Riccardo and Zakharyaschev, Michael},
  year = {2018},
  month = jul,
  pages = {5511--5519}
}

@inproceedings{KaplanKeimSchneider2024_1000171637,
  title = {Combining Knowledge Graphs and Large Language Models to Ease Knowledge Access in Software Architecture Research},
  booktitle = {Proceedings of the 2nd International Workshop on Semantic Technologies and Deep Learning Models for Scientific},
  author = {Kaplan, Angelika and Keim, Jan and Schneider, Marco and Koziolek, Anne and Reussner, Ralf},
  year = {2024},
  series = {{{CEUR}} Workshop Proceedings},
  volume = {3697},
  pages = {76--82}
}

@inproceedings{Agarwal2024,
  title = {Bring Your Own {{KG}}: {{Self-supervised}} Program Synthesis for Zero-Shot {{KGQA}}},
  booktitle = {Findings of the Association for Computational Linguistics: {{NAACL}} 2024},
  author = {Agarwal, Shiyue and Menon, Rishab and Singh, Sameer and Gardner, Matt and Khashabi, Daniel},
  year = {2024},
  pages = {837--859}
}

@inproceedings{liao2020questioning,
  title = {Questioning the AI: informing design practices for explainable AI user experiences},
  author = {Liao, Q Vera and Gruen, Daniel and Miller, Sarah},
  booktitle = {Proceedings of the 2020 CHI conference on human factors in computing systems},
  pages = {1--15},
  year = {2020}
}

@inproceedings{Ovadia2024,
  title = {Fine-Tuning or Retrieval? {{Comparing}} Knowledge Injection in Llms},
  booktitle = {Proceedings of the 2024 Conference on Empirical Methods in Natural Language Processing ({{EMNLP}})},
  author = {Ovadia, Oded and Brief, Menachem and Mishaeli, Moshik and Elisha, Oren},
  year = {2024},
  month = jan,
  pages = {237--250}
}

@inproceedings{nguyen2023advancing,
  title = {Advancing Data Mining Results Sharing with {{ML-schema}}},
  booktitle = {Proceedings of the International Conference on Data Mining},
  author = {Nguyen, P. and Barrett, D.},
  year = {2023},
  pages = {300--315}
}

@misc{openai_chat_format,
  title = {{{OpenAI}} Chat Completion {{API}} Format},
  author = {{OpenAI}},
  year = {2024}
}

@misc{doshivelez2017rigorousscienceinterpretablemachine,
  title = {Towards a Rigorous Science of Interpretable Machine Learning},
  author = {{Doshi-Velez}, Finale and Kim, Been},
  year = {2017},
  primaryclass = {stat.ML},
  archiveprefix = {arXiv}
}

@misc{edge2025localglobalgraphrag,
  title = {From Local to Global: A Graph {{RAG}} Approach to Query-Focused Summarization},
  author = {Edge, Darren and Trinh, Ha and Cheng, Newman and Bradley, Joshua and Chao, Alex and Mody, Apurva and Truitt, Steven and Metropolitansky, Dasha and Ness, Robert Osazuwa and Larson, Jonathan},
  year = {2025},
  primaryclass = {cs.CL},
  archiveprefix = {arXiv}
}

@article{SCHUCHTER202561,
title = {Application of artificial intelligence in model-based systems engineering of automated production systems},
journal = {Procedia CIRP},
volume = {136},
pages = {61-66},
year = {2025},
issn = {2212-8271},
author = {Timo Schuchter and Patrick Saft and Ralf Stetter and Markus Pfeil and Wolfram Höpken and Markus Till and Stephan Rudolph}
}

\end{document}